\pdfoutput=1

\documentclass[11pt]{article}

\usepackage[]{EMNLP2022}

\usepackage{times}
\usepackage{latexsym}

\usepackage[T1]{fontenc}

\usepackage[utf8]{inputenc}

\usepackage{microtype}

\usepackage{inconsolata}

\usepackage[ruled,vlined]{algorithm2e}
\usepackage{graphicx}
\usepackage{booktabs}
\usepackage{multirow}
\usepackage{pifont}
\newcommand{\cmark}{\ding{51}}%
\newcommand{\xmark}{\ding{55}}%
\usepackage{subcaption}
\usepackage{caption}

\newcommand{\daniella}[1]{}
\newcommand{\qi}[1]{}
\newcommand{\tao}[1]{}
\newcommand*\samethanks[1][\value{footnote}]{\footnotemark[#1]}

\title{Augmenting Multi-Turn Text-to-SQL Datasets with Self-Play}

\author{Qi Liu$^{1}$\thanks{~~~Equal Contribution}, Zihuiwen Ye$^{2}$\samethanks, Tao Yu$^{1}$, Phil Blunsom$^{2}$, Linfeng Song$^{3}$ \\
$^{1}$The University of Hong Kong,  $^{2}$University of Oxford,  $^{3}$Tencent AI Lab, Bellevue, WA, USA \\
\texttt{\{liuqi, tyu\}@cs.hku.hk}; \texttt{\{zihuiwen.ye, phil.blunsom\}@cs.ox.ac.uk}; \\\texttt{lfsong@tencent.com}}

\begin{document}
\maketitle
\begin{abstract}
The task of context-dependent text-to-SQL aims to convert multi-turn user utterances to formal SQL queries. This is a challenging task due to both the scarcity of training data from which to learn complex contextual dependencies and to generalize to unseen databases. In this paper we explore augmenting the training datasets using self-play, which leverages contextual information to synthesize new interactions to adapt the model to new databases. We first design a SQL-to-text model conditioned on a sampled goal query, which represents a user’s intent, that then converses with a text-to-SQL semantic parser to generate new interactions. We then filter the synthesized interactions and retrain the models with the augmented data.
We find that self-play improves the accuracy of a strong baseline on SParC and CoSQL, two widely used cross-domain text-to-SQL datasets. Our analysis shows that self-play simulates various conversational thematic relations,  enhances cross-domain generalization and improves beam-search.\footnote{Our code is available at: \url{https://github.com/leuchine/self_play_picard}} 
\end{abstract}


\section{Introduction}
Multi-turn text-to-SQL translation is a powerful semantic parsing paradigm that converts natural language user utterances into executable SQL queries in a conversational environment. 
Compared to regular text-to-SQL tasks such as Spider \cite{yu2018spider} and GeoQuery \cite{Zelle1996LearningTP}, conversational text-to-SQL requires interpreting coreference and omission phenomena that frequently appear in human conversations. 
To be effective, text-to-SQL models must uncover complex contextual dependencies while grounding user utterances in task specific database schemas. 

Numerous architectures and pretraining methods have been proposed for tackling context-dependent text-to-SQL \cite{suhr-etal-2018-learning,zhang-etal-2019-editing,hui-2021-dynamic,scholak-2021,yu2021score,xie-2022-skg}.
However, the size of the datasets used has been limited due to the high cost of annotating multi-turn dialogue and SQL pairs, which often requires trained experts. 
Existing multi-turn text-to-SQL datasets, such as SParC \cite{yu2019sparc} and CoSQL \cite{yu2019cosql}, require text-to-SQL parsers to generalize to unseen databases at test time, but doing so is difficult with limited training context. 



\begin{figure*}[!htbp]
 \centering
 \includegraphics[width=0.95\textwidth]{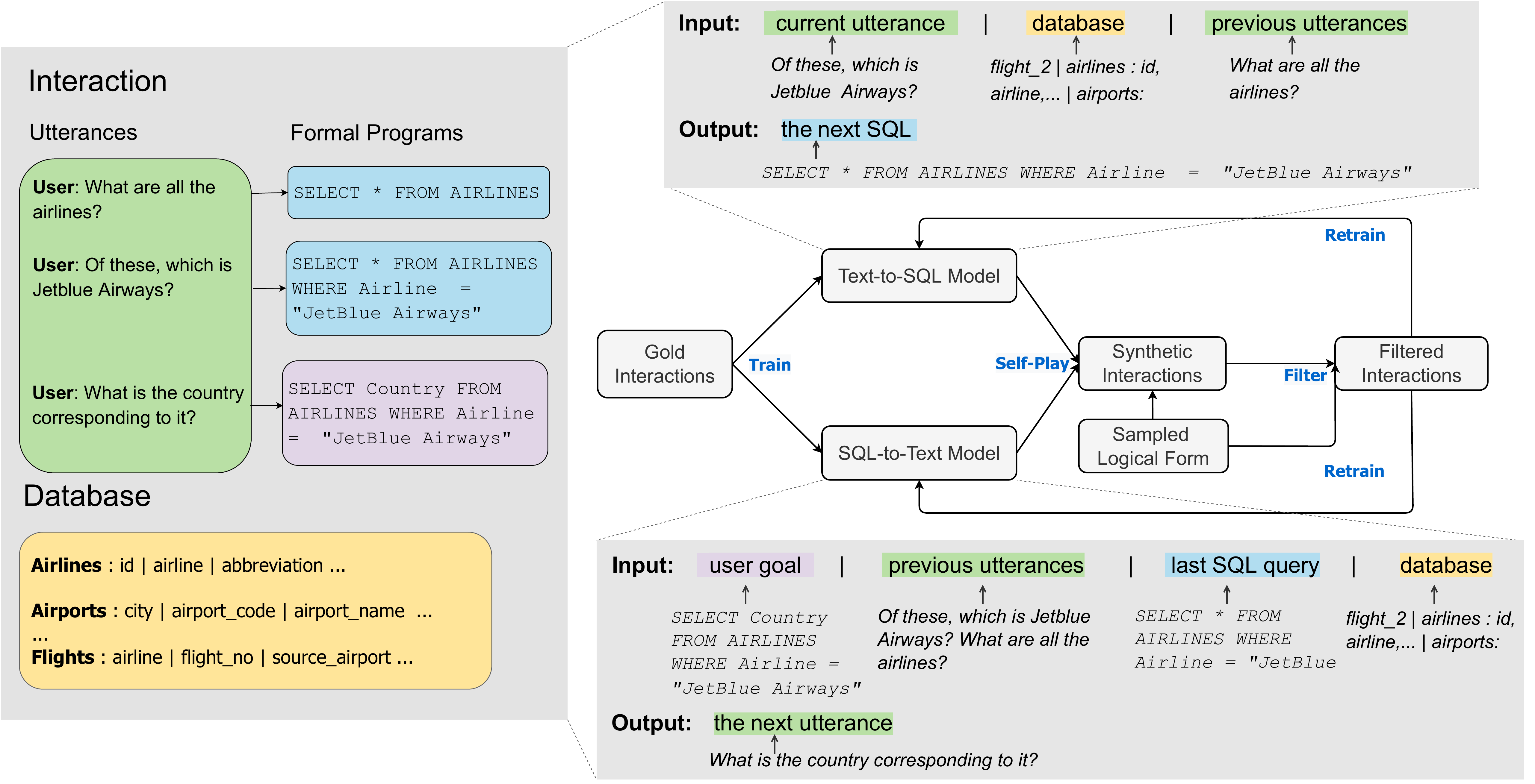}
 \caption{Multi-turn text-to-SQL with self-play. We transform an interaction from SParC on the left to seq2seq formats (top: text-to-SQL,  bottom: SQL-to-text). User utterances, SQL queries, databases, and user goals are concatenated  with a `` | '' symbol and shown in green, blue, yellow, and purple respectively. We use self-play to generate synthetic interactions. The synthetic interactions are filtered and used to retrain the text-to-SQL and SQL-to-text models. 
 \label{fig:overview}}
\end{figure*}

In this paper we propose the use of self-play to augment multi-turn text-to-SQL datasets in order to achieve more robust generalization. Self-play simulates interactions between multiple artificial agents in order to generate a training signal in addition to supervised data. It has been successfully applied in a wide range of tasks, e.g.\ board games \cite{silver2016mastering, silver2018general} and multiplayer battle games \cite{vinyals2019grandmaster, berner2019dota}. It has also been applied in dialogue simulations, during which a dialogue model converses with a user simulator to generate synthetic dialogues \cite{schatzmann2006survey, gur2018user, tseng-etal-2021-transferable}. In our work, we extend self-play to semantic parsing. 

 Although self-play has been adopted in task-oriented dialogue, the need to pre-define a domain specific ontology of slot-value pairs (e.g.\ the slot value ``price=expensive'' for a restaurant booking) \cite{henderson-etal-2014-second, wen-2016, Budzianowski-2018} prevents self-play from simulating interactions in a new domain.
Adding a new domain for task-oriented dialogue is difficult and labor-intensive. On the other hand, text-to-SQL tasks \cite{yu2018spider, yu2019sparc, yu2019cosql} use a domain-independent formalism, i.e.\ SQL queries. We demonstrate that self-play is well-suited to simulating interactions in a new domain given a database schema, improving cross-domain generalization.

We use PICARD \cite{scholak-2021} as the base of our text-to-SQL model. When generating a new interaction, we first sample a SQL query with \citet{zhong2021grounded} as the goal query and condition the SQL-to-text model on this sampled SQL. The text-to-SQL model converses with the SQL-to-text model to simulate a new interaction. We filter out the interactions that are not grounded to the sampled goals and employ self-training \cite{yarowsky1995unsupervised, zoph2020rethinking} to retrain the text-to-SQL model and the SQL-to-text model. We conduct extensive experiments on SParC and CoSQL. Our main findings are:  

\begin{itemize}
    \itemsep0em
    \item Self-play helps the text-to-SQL model learn various conversational thematic relations (\S\ref{sec:case_analysis}) and improves cross-domain generalization (\S\ref{sec:template_analysis}).
    
    \item Self-play improves the performance on the majority of SQL types. Models after self-play perform particularly well on queries of medium difficulty (\S\ref{sec:template_analysis}).
    
    \item Self-play improves beam search. Models after self-play are less sensitive to the beam size and can perform well with even small beam sizes (\S\ref{sec:beam_analysis}). 
    
\end{itemize}

\section{Preliminary}
In this section, we formally define the multi-turn text-to-SQL task and introduce the PICARD \cite{scholak-2021} model, which we use as our baseline. PICARD obtains state-of-the-art results on several text-to-SQL tasks. 

\subsection{Task Definition} 
In context-dependent text-to-SQL tasks, we are given interactions between a user and a system. Each interaction spans multiple turns. The user ends the interaction when the query returns the required information from the database. Formally, at each turn $t$ (where $1 \leq t\leq T$), multi-turn text-to-SQL produces a valid and executable SQL query $\mathcal{Q}_t$ given a database $\mathcal{D}$, a current user utterance $\mathcal{U}_t$, and a dialogue context $\mathcal{C}_t$ (which is usually the previous user utterances $\mathcal{U}_{<t}$):

\begin{equation}
    \label{eq.text2seql_model}
    p(\mathcal{Q}_t \mid \mathcal{U}_t, \mathcal{C}_t, \mathcal{D}).
\end{equation}







\subsection{Baseline: PICARD}



We use PICARD \cite{scholak-2021} as our baseline conditional model for Equation \ref{eq.text2seql_model}. PICARD serializes the database schema $\mathcal{D}$ into a sequence following \citet{lin-etal-2020-bridging}. An example of the input and output format is shown in Figure \ref{fig:overview}. 
PICARD finetunes T5 \cite{raffel-2019-t5}, a sequence-to-sequence transformer, with input and output sequences. PICARD proposes an incremental parsing method for constrained decoding during beam search. Specifically, it rejects inadmissible tokens at each beam search step subject to parsing rules that encode lexical and grammatical constraints. Only the beam hypotheses that pass all the constraint checks are kept. PICARD also leverages SQL schema information, such as the column names of each table, to impose checks on the validity of the generated SQL. PICARD greatly reduces the likelihood of decoding invalid SQL queries.

\section{Method}
Here we introduce how we use self-play for data augmentation. We first design a SQL-to-text model (\S\ref{sec:sql2text}). Next, we describe how to use self-play to generate synthetic interactions (\S\ref{sec:self-play}). Finally, we explain how we incorporate the generated data for self-training (\S\ref{sec:self-train}). 

\subsection{The SQL-to-Text Model \label{sec:sql2text}}
We design a user simulator, which is a SQL-to-text model, to converse with the text-to-SQL model to generate synthetic interactions. Specifically, at each turn $t$ we would like the user simulator to produce a meaningful question that would naturally be asked by a human user. In each interaction, a user has a goal to achieve. We explicitly condition the SQL-to-text model on a user goal, $\mathcal{G}$, to encourage the user simulator to ask questions that are grounded to this goal. Formally, the SQL-to-text model calculates the following conditional at each turn:
\begin{equation}
    \label{eq.seql2text_model}
    p(\mathcal{U}_t \mid \mathcal{Q}_{t-1}, \mathcal{C}_{t}, \mathcal{G}, \mathcal{D}),
\end{equation}
where the context $\mathcal{C}_t$ contains the previous user utterances $\mathcal{U}_{<t}$. During training, $\mathcal{G}$ is the SQL query of the final turn $T$, i.e.\ $\mathcal{Q}_T$.
During inference we adopt \citet{zhong2021grounded} to sample a new goal query as shown in \S\ref{sec:self-play}. 
We employ the seq2seq approach and parameterize the SQL-to-text model (Eq. \ref{eq.seql2text_model}) with T5. We concatenate the user goal $\mathcal{G}$, the last SQL query $\mathcal{Q}_{t-1}$, the previous user utterances $\mathcal{U}_{<t}$, and the serialized schema $\mathcal{D}$ to predict the next user utterance $\mathcal{U}_t$. For example, one input would be: ``user goal $\mid$ previous utterances $\mid$ last SQL query $\mid$ serialized database''. Its target label is the correct user utterance for the next turn. We pad the last utterance with a special stop-of-interaction symbol. In SQL-to-text, there could be multiple reasonable questions to ask for the next turn, i.e.\ a one-to-many relation. A well-trained SQL-to-text model can generate new questions, thereby increasing the diversity of user dialogue flows in the dataset and improving generalization.

\begin{algorithm}[t]
\small
\SetAlgoNoEnd 
\SetAlgoLined
\SetKwInOut{Input}{Input}
\SetKwInOut{Output}{Output}
\SetKw{Continue}{continue}
\SetKw{Or}{or}
\SetKw{In}{in}
\Input{Gold interactions $\mathcal{I}$, \# iteration $k$ for synthetic data generation, threashold $w$.}
\Output{A text-to-SQL model and a SQL-to-text model.}
 Pretrain a text-to-SQL model  $p(\mathcal{Q}_t | \mathcal{U}_t, \mathcal{C}_t, \mathcal{D})$ and a SQL-to-text model $p(\mathcal{U}_t | \mathcal{Q}_{t-1}, \mathcal{C}_{t}, \mathcal{G}, \mathcal{D})$ on $\mathcal{I}$.\\
 $\mathcal{I}' = \emptyset$ \\ 
\For{$i$ \In $(1, ..., k)$}{
 Sample a goal query $\mathcal{G}$. \\ 
 Generate a synthetic interaction $\mathcal{I}_S$ by self-play between text-to-SQL and SQL-to-text.\\
 Calculate $\textit{score}(\mathcal{Q}_T, \mathcal{G})$ on $\mathcal{I}_S$. \\
 \If{$\textit{score}(\mathcal{Q}_T, \mathcal{G}) > w$}{Add $\mathcal{I}_S$ to $\mathcal{I}'$}
 }
 Retrain $p(\mathcal{Q}_t | \mathcal{U}_t, \mathcal{C}_t, \mathcal{D})$ and $p(\mathcal{U}_t | \mathcal{Q}_{t-1}, \mathcal{C}_{t}, \mathcal{G}, \mathcal{D})$ on $\mathcal{I} \cup \mathcal{I}'$.
 
 \Return the retrained text-to-SQL model and the SQL-to-text model. 
 \caption{Self-play for text-to-SQL}
\caption{Self-training.\label{al:self_training}}
\end{algorithm}

 

\subsection{Self-Play\label{sec:self-play}}
We pretrain both the text-to-SQL and SQL-to-text models on the gold training data by minimizing the negative log likelihood: 

\begin{equation}
    \label{eq.nll}
     L = -\sum^N_{i=1}\sum^V_{j=1} \log{p(y^i_j \mid y^i_1, y^i_2, \ldots, y^i_{j-1})},
\end{equation}
where $N$ is the number of training examples, $V$ is the sequence length, and each $y^i_j$ is a token in the reference sequence. With the models pretrained on the gold dialogues, we can generate synthetic interactions using self-play. First, we need to specify a SQL query as the eventual goal $\mathcal{G}$ of the interaction. We adopt the query sampling method proposed in \citet{zhong2021grounded} for synthesizing a goal $\mathcal{G}$. \citet{zhong2021grounded} first builds and samples coarse SQL templates with the SQLs in the training set by replacing the column and value mentions in the queries with typed slots. For example, \texttt{SELECT T1.id, T2.name} is converted to the template \texttt{SELECT key1, text1}. To adapt the models to an unseen environment, they sample an unseen database and fill in the typed slots with columns and values from the sampled database to form a new SQL query. We follow this approach to synthesize goals in new domains for cross-domain generalization. The complete sampling procedure is given in Appendix \ref{app:query_sampling}. We concatenate the sampled goal $\mathcal{G}$ with an empty context and the serialized schema as shown in Eq. \ref{eq.seql2text_model} and feed it into the SQL-to-text model to produce the first user utterance. Then, the text-to-SQL model and SQL-to-text
model can continue the interaction with Eq. \ref{eq.text2seql_model} and Eq. \ref{eq.seql2text_model} until the end. A synthetic interaction ends whenever the SQL-to-text model decodes the stop-of-interaction symbol.

\begin{table*}[!tbp]
\centering
\scalebox{0.8}{
\begin{tabular}{lllllllll}
\hline
\textbf{Dataset} &\textbf{System Response} & \textbf{\# Dialogues} & \textbf{\#Turns } & \textbf{\# Databases} & \textbf{\# Domains} & \textbf{\# Tables} & \textbf{Avg. Q len} & \textbf{Vocab} \\
\hline
SParC & \xmark & 4,298 & 12,726  & 200 & 138 & 1,020  & 8.1 & 3,794  \\
CoSQL & \cmark & 3,007 & 15,598  & 200 & 138 & 1,020  & 11.2 & 9,585 \\
\hline
\end{tabular}
}
\caption{Comparison of cross-domain context-dependent text-to-SQL datasets.  \label{tab:datasets_stats}
}
\end{table*}

\paragraph{Filtering} Synthetic conversations generated by self-play may diverge from the sampled goals. To filter these low-quality conversations, we compare the generated SQL query $\mathcal{Q}_T$ from the last turn $T$ with the sampled goal $\mathcal{G}$ (see \S\ref{sec:self-play}) using a similarity score $\textit{score}(\mathcal{Q}_T, \mathcal{G})$. We follow \citet{yu2018spider} and decompose the SQL queries $\mathcal{Q}_T$ and $\mathcal{G}$ into SQL substructures $\mathcal{Q}_{T_{s}}, \mathcal{G}_{s} $ (e.g. \texttt{select, where, group\_by, order\_by} statements) and calculate the accuracy on each substructure.
We let $\textit{score}(\mathcal{Q}_T, \mathcal{G})$ be the average of the accuracy over all the substructures. We keep a synthetic conversation if $\textit{score}(\mathcal{Q}_T, \mathcal{G})$ is larger than a threshold value $w$. A high score means that the synthetic conversation is grounded to the sampled goal.

\subsection{Self-Training\label{sec:self-train}}

We re-train a new text-to-SQL model and a new SQL-to-text model with both the gold training data and the filtered synthetic interactions. Algorithm \ref{al:self_training} shows the overall procedures. The complete self-play and self-training steps are shown in Figure \ref{fig:overview}. Our method is an instance of self-training as the models are re-trained with their own outputs. To retrain the text-to-SQL and SQL-to-text models, we can either combine the filtered synthetic data with gold interactions, or pretrain on the synthetic interactions before fine-tuning on the gold interactions. We employ the second approach as we observe that the second approach performs slightly better than the first one. 
\tao{in principle you could self-play multiple times?} \daniella{Yes, we could self-play multiple times which will likely improve the result. We only ran it once due to computational constraints. We are thinking of leaving that to the future work (limitations section)}

\section{Datasets and Main Results} \label{datasets}
In this section, we evaluate the performance of self-play on cross-domain multi-turn semantic parsing. We first introduce the datasets (\S\ref{sec:datasets}), then detail the evaluation metrics (\S\ref{sec:settings}), and finally we show the main results (\S\ref{sec:results}).

\subsection{Datasets\label{sec:datasets}}
\tao{maybe this part can be shortened if the page number exceeds}
We evaluate our method on two large-scale benchmark datasets, SParC \cite{yu2019sparc} and CoSQL \cite{yu2019cosql}. Table \ref{tab:datasets_stats} summarises the statistics of the two datasets. Following PICARD, we additionally pretrain the text-to-SQL model on a single-turn text-to-SQL dataset Spider \cite{yu2018spider}.
\tao{so the text-to-SQL model is also trained on Spider?} \daniella{Yes, maybe I should write more explicitly. Is it better to have a separate paragraph to introduce Spider?}
All these datasets require generalization to new domains as they contain different databases for training, development, and testing, respectively, to evaluate the cross-domain performance. We discuss SParC and CoSQL in detail. 


\paragraph{SParC} SParC is a multi-turn text-to-SQL dataset that spans 200 databases in which the tables cover 138 different domains. Each question in an interaction belongs to one of the four thematic relations: refinement, theme-entity, theme-property, and answer refinement \cite{bertomeu-2006-contextual}. For example, given a question ``Which major has the fewest students?", the next query can be an ``refinement'' query, ``What is the most popular one?", which asks for the same entity as the previous question but with a different constraint. 

\paragraph{CoSQL} CoSQL is the dialogue version of SParC. In CoSQL, besides a SQL query, the system also generates a natural language response. It is collected with the Wizard-of-Oz setting \cite{Budzianowski-2018}. The dataset is used for three tasks including state-tracking, user act prediction, and response generation. We use this dataset for state-tracking, where the goal is to map user utterances into a SQL query at each turn.

\begin{table*}[!thbp]
	\centering
	\scalebox{0.8}{
	\begin{tabular}{cc cccc cccc}
		\toprule 
		{\multirow{3}*{}} & {\multirow{3}*{}} & \multicolumn{4}{c}{SParC} & \multicolumn{4}{c}{CoSQL} \\
		\cmidrule{3-10}
		\multirow{2}{*}{} & \multirow{2}{*}{\textbf{Models}} &
		\multicolumn{2}{c}{Dev} &
		\multicolumn{2}{c}{Test} &
		\multicolumn{2}{c}{Dev} &
		\multicolumn{2}{c}{Test} \\
		\cmidrule{3-10}
		& & QM & IM & QM & IM & QM & IM & QM & IM  \\
		\midrule
		{\multirow{4}*{T5-Base}} &  w/ PICARD w/ self-play & \textbf{62.4} & \textbf{42.1} &  - & - & \textbf{53.0} & \textbf{21.5} & - & - \\
	                             & 	w/ PICARD w/o self-play & 57.5 & 38.6 &  - & - & 49.9 & 20.2 & - & - \\
	                             & w/o PICARD w/ self-play & 57.2 & 37.5 &  - & - & 50.6 & 20.4 & - & - \\
                                 & w/o PICARD w/o self-play & 50.3 & 31.7 &  - & - & 45.2 & 18.7 & - & - \\
        \midrule
        
        \multirow{4}{*}{T5-Large} & w/ PICARD w/ self-play & \textbf{65.5} & \textbf{45.6} &  \textbf{64.0} & \textbf{39.6} & \textbf{55.7} & \textbf{23.2} & \textbf{53.4} & \textbf{22.7} \\
                                 & w/ PICARD w/o self-play & 63.0 & 43.0 &  60.7 & 36.9 & 54.3 & 21.9 & 52.1 & 21.6 \\
                                 & w/o PICARD w/ self-play & 64.1 & 44.1 &  - & - & 53.9 & 21.2 & - & - \\
                                 & w/o PICARD w/o self-play & 57.5 & 38.1 &  - & - & 51.4 & 20.6 & - & - \\
		
		\bottomrule
	\end{tabular}
	}
	\caption{Main results. Models after self-play outperform the baselines under different configurations.  \label{tab:main_results}}
\end{table*}





\subsection{Settings and Evaluation Metrics\label{sec:settings}}
Following \citet{yu2018spider}, we measure the performance with \textit{question match} (QM) and \textit{interaction match} (IM), both of which are based on the exact set match accuracy. The exact set match is computed by decomposing the predicted SQLs into clauses such as \texttt{SELECT, WHERE, GROUP BY} and calculating the set matching score on each. QM is 1 if the exact set match for a question in an interaction is 1. IM is 1 if the exact set matches for all questions in an interaction are 1. The number of the self-play generated training data for SParC (CoSQL) before filtering is 100,000 (100,000) and 49,623 (48,291) after filtering. Appendix \ref{app:implementation} shows implementation details of our experiments. 


\subsection{Main Results\label{sec:results}}

We report the main results in Table \ref{tab:main_results}. We observe that the configuration ``w/ PICARD w/ self-play'' achieves the best results on both datasets (measured by QM and IM).
This demonstrates the benefit of self-play. The improvement brought by self-play is more salient on SParC than on CoSQL, while T5-Large w/ PICARD w/ self-play outperforms the vanilla T5-3B reported by \citet{scholak-2021}.
Therefore, we conclude that self-play is an effective data augmentation method to improve performance on cross-domain context-dependent text-to-SQL. Appendix \ref{app:analysis} shows the system's performance under different configurations of the generated synthetic data.

\section{Analysis}

In this section we take SParC and systematically analyze the effect of self-play. First, to gain more insight into how a question's position or the query template affect the models, we examine self-play performance stratified by different turn number and SQL templates in \S\ref{sec:template_analysis}. Then, we study whether self-play improves decoding during beam search (\S\ref{sec:beam_analysis}). We further conduct a case study of self-play interactions in \S\ref{sec:case_analysis}.

\subsection{Turn and Template Analysis\label{sec:template_analysis}}

\begin{figure}[t]
 \centering \includegraphics[width=0.80\linewidth]{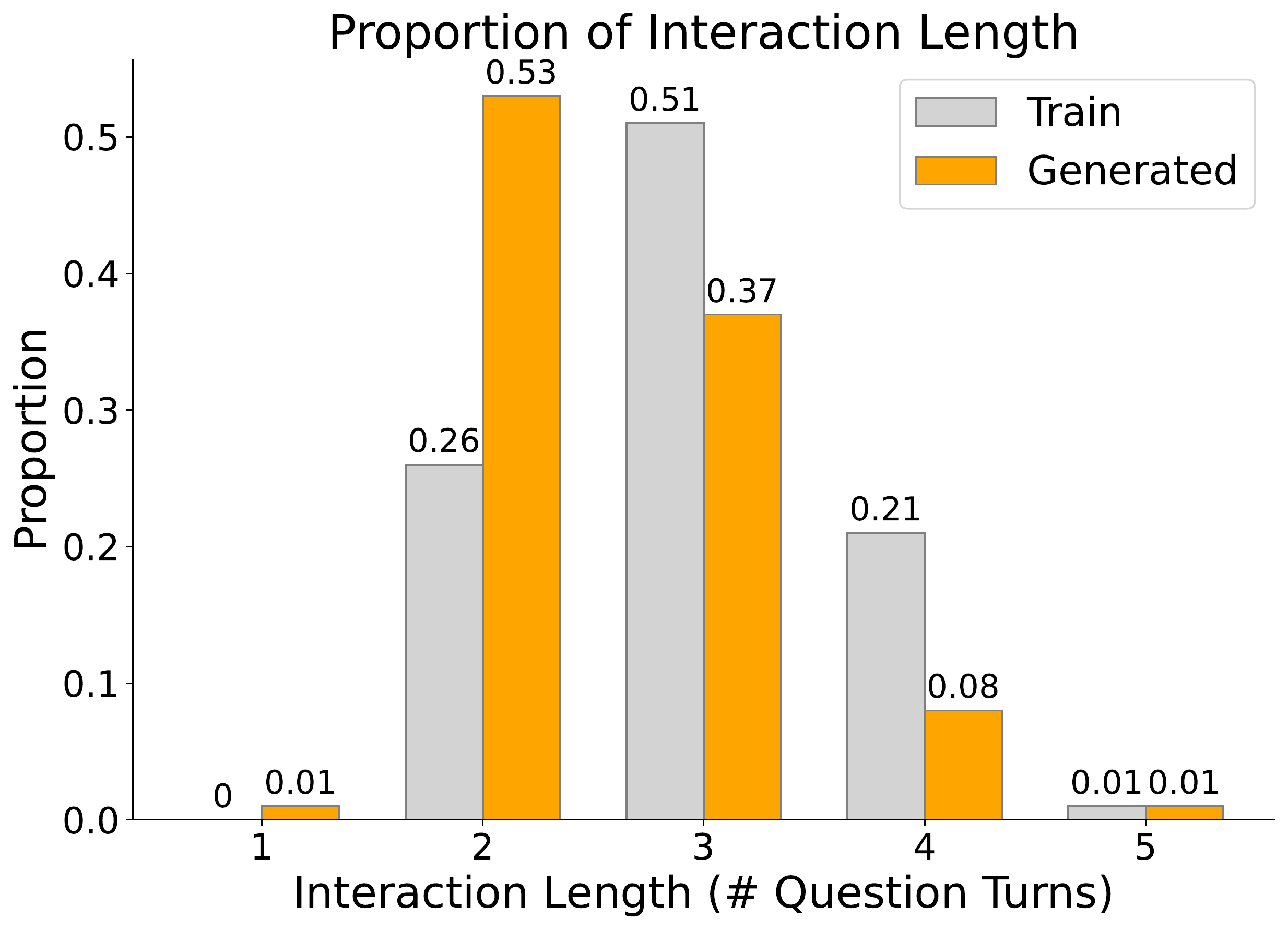}
 \vspace*{-5pt}
 \caption{The distribution of interaction lengths of gold interactions and self-play interactions. \label{fig:lengths}}
\end{figure}

\begin{figure}[t]
 \centering
 \includegraphics[width=0.76\linewidth]{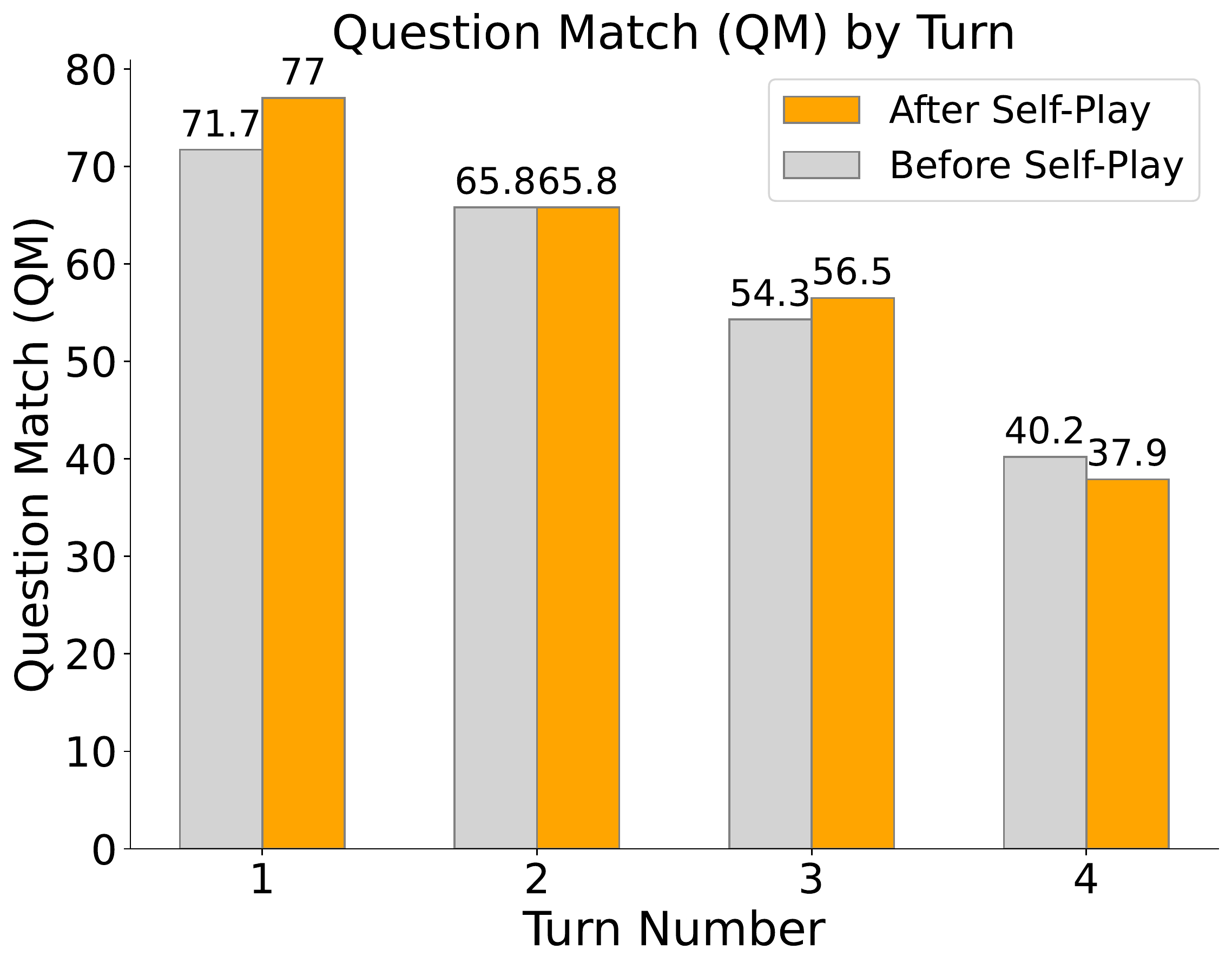}
 \vspace*{-5pt}
 \caption{\textit{Question Match} (QM) by turn numbers.}
 \label{fig:by_turn}
\end{figure}

We first plot the distribution of interaction lengths in Figure \ref{fig:lengths}. Self-play produces shorter interactions with a mean length of 2.53, whereas the mean of the training data is 2.97. 
Figure \ref{fig:by_turn} shows \textit{Question Match} (QM) accuracy stratified by question turns. 
The performance after self-play increases on the turn numbers $\leq3$ and decreases on the turn number 4. This is because self-play does not generate enough long interactions as shown in Figure \ref{fig:lengths}. 

\begin{figure}[t]
 \centering
 \includegraphics[width=0.35\textwidth]{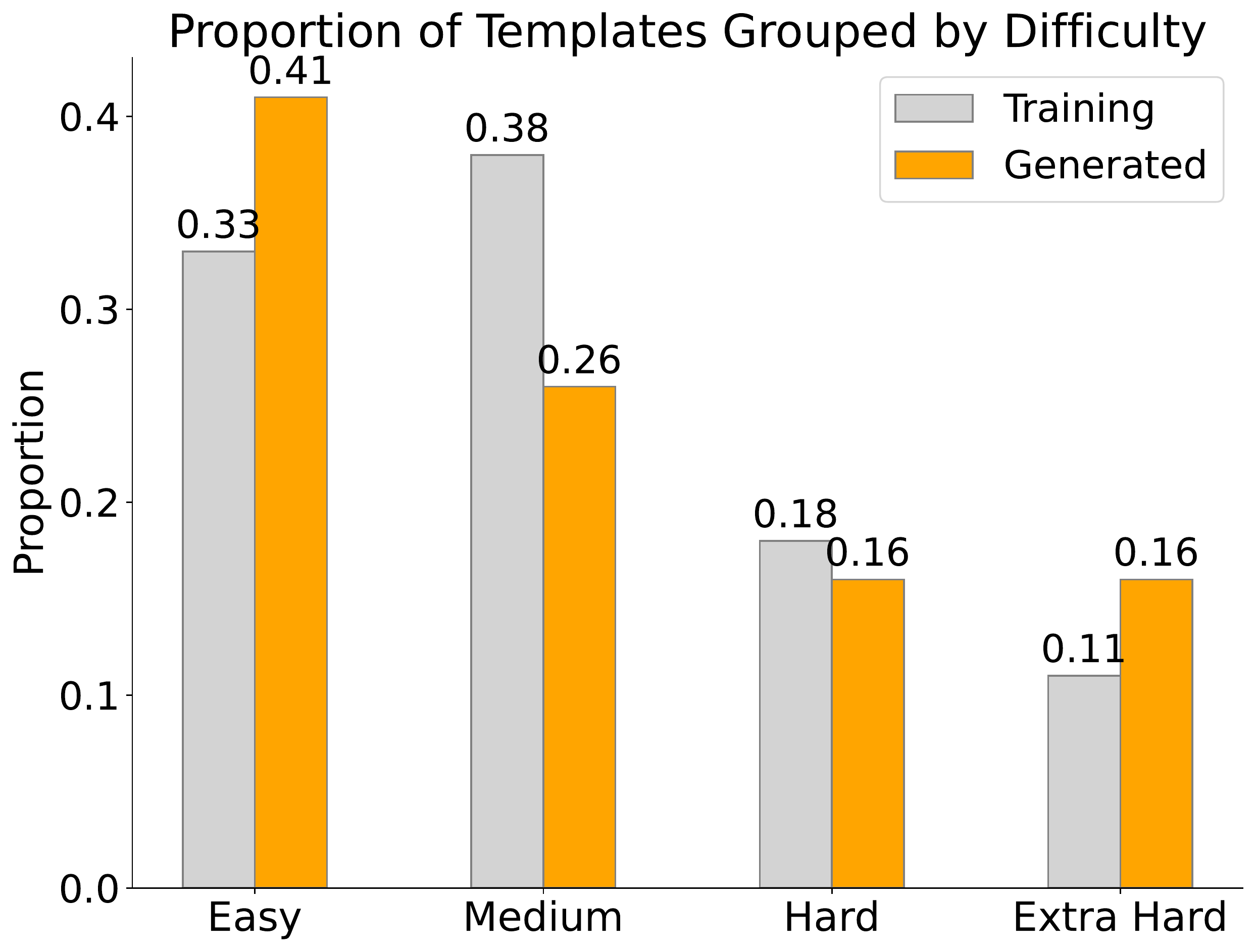}
 \caption{The distribution of template difficulties.
 \label{fig:difficulty}}
\end{figure}

Next we compare the performance of the models with and without self-play stratified by the difficulty of the SQL template. We first convert SQLs into templates using the method in \citet{zhong2021grounded}. To get a sense of the overlap of the templates in self-play and training, 85\% of self-play templates occur in the training templates. That is to say, 15\% of self-play templates are new templates that are unseen during training. As shown in Figure \ref{fig:difficulty}, self-play interactions have higher proportions of easy and extra hard templates and lower proportions of medium and hard templates. We compare the performance before and after self-play on the SParC validation set in Table \ref{tab:difficulty}. Self-play brings the largest improvement to interactions of medium difficulty, followed by hard and easy ones.

On manually inspecting the performance for templates we observe that the performance on most is improved after self-play. Of the 72 unique templates in the SParC validation set, there are only 12 query templates whose performance decreases. The performance on the templates with the operator ``\texttt{select counts}'' improves significantly (on average an increase of 12 for the 11 templates with ``\texttt{select counts}''), possibly because the word ``count'' appears more often in the generated ones than training. For example, ``\texttt{select count (*\_col\_0)}'' is one of the top templates in the generated dataset as shown in Table \ref{tab:top_templates}. We find that the accuracy of the template ``\texttt{select sum (number\_col\_0)}'' increases from 50 to 100 after self-play. Self-play also reduces hallucination to some extent. For example, when asked to display certain record companies, the model would hallucinate the constraint ``\texttt{having count(*)>2}'' before self-play, but the system gives the correct result after self-play. These results confirm the effectiveness of self-play. Appendix \ref{app:more_templates} shows more examples of generated templates and the improvement brought by self-play. 
\tao{maybe consider to make this part shorter if the page number exceeds}

\begin{table}[!h]
\centering
\scalebox{0.7}{
\begin{tabular}{lllll}
\hline
\textbf{Difficulty}  & \textbf{Before self-play} & \textbf{After self-play} & \textbf{Improvement}\\
\hline
Easy  & 80.9 & 82.4  & 1.5  \\
Medium  & 62.6 & 67.7  & 5.1\\
Hard  & 41.2  & 44.5 & 3.4 \\
Extra Hard & 31.8 & 31.8 & 0 \\
\hline
\end{tabular}}
\caption{The performance before and after self-play on SParC validation set grouped by template difficulties. 
\label{tab:difficulty}
}
\end{table}


\begin{table*}[t]
\centering
\scalebox{0.85}{
\begin{tabular}{llll}
\hline
\textbf{Top Templates in Train} & \textbf{Proportion} & \textbf{Top Templates in Self-Play} & \textbf{Proportion} \\
\hline
\texttt{select text\_col\_0} &  7.56\% & \texttt{select text\_col\_0} & 13.33\% \\
\texttt{select text\_col\_0 where text\_col\_1 = value} & 4.99\% & \texttt{select *\_col\_0}  & 5.13\% \\
\texttt{select *\_col\_0} & 3.67\% & \texttt{select count ( *\_col\_0 )} & 4.57\% \\
\hline
\end{tabular}}
\caption{The top templates and their proportions in the SParC training and generated data. \label{tab:top_templates}}
\end{table*}

\begin{table*}[h]
    \begin{subtable}[]{1\textwidth}
        \centering
        \scalebox{0.72}{
        \begin{tabular}{llll}
        
            \toprule
            \textbf{Turn} & \textbf{User Utterance} & \textbf{System Query}       &  \textbf{Execution Result} \\
            \midrule
            \multirow{2}{*}{1}
             &  Show the location for all gas stations. & \texttt{select location} & Herne Hill, Channel Hill  \\ 
             &                                   & \texttt{from gas\_station}      & Reading North ...\\ 
            \midrule 
            \multirow{4}{*}{2}
             &                                     &  \texttt{select t3.location, t2.company from}  &   \\
             & For each of them, also show the    &  \texttt{station\_company as t1 join company}   &  (Hennry Hill, Petrobras) \\
             & company names.                   &  \texttt{as t2 on t1.company\_id = t2.company\_id} & (Maindy Ave, BP)... \\
             &                                  &  \texttt{join gas\_station as t3 on}               & \\
             &                                  &  \texttt{t1.station\_id = t3.station\_id}          & \\
            \midrule
            \multirow{4}{*}{3}
            &                                         & \texttt{select t3.location, t2.company from} &  (Herne Hill, Royal Dutch Shell)\\
            & Order them by assets in descending order.   & \texttt{station\_company as t1 join company} & (Herne St, ExxonMobil)... \\
            &                                        & \texttt{as t2 on t1.company\_id = t2.company\_id} &\\
            &                                        & \texttt{join gas\_station as t3 on} & \\
            &                                        & \texttt{t1.station\_id = t3.station\_id order} & \\
            &                                        & \texttt{by t2.assets\_billion desc} &\\
            
            \bottomrule
        \end{tabular}
        }
        \captionsetup{justification=centering}
        \caption{\textit{A successful interaction.}}
        \label{tab:sub_a}
    \end{subtable}\vspace{5mm}
     
    \begin{subtable}[]{\textwidth}
        \centering
        \scalebox{0.78}{
        \begin{tabular}{llll}
        \toprule
        \textbf{Turn} & \textbf{User Utterance} & \textbf{System Query}       &  \textbf{Execution Result} \\
        \midrule
        \multirow{2}{*}{1}
         &  Show the names of all documents. & \texttt{select document\_name} & Introduction of OS, \\ 
         &                                   & \texttt{from documents}        & How to write a CV... \\ 
        \midrule 
        \multirow{2}{*}{2}
        & For each of them, also show the  & \texttt{select document\_name, count(*) from} & (A history of Arts, 1),  \\
        & number of templates.             & \texttt{documents group by document\_name} &  (About Korea, 1)...        \\
        \midrule
        \multirow{3}{*}{3}
        & Order the document names in descending & \texttt{select document\_name from documents} & Welcome to NY,  \\
        & order of the number of templates.      & \texttt{group by document\_name order} &  Understanding DB... \\
        &                                        & \texttt{by count(*) desc} & \\
        \midrule
        \end{tabular}}
        \caption{\textit{Failure case: language drift.}}\vspace{5mm}
        \label{tab:sub_b}
    \end{subtable}
    
    \begin{subtable}[]{\textwidth}
        \centering
        \scalebox{0.8}{
        \begin{tabular}{llll}
        \toprule
        \textbf{Turn} & \textbf{User Utterance} & \textbf{System Query}       &  \textbf{Execution Result} \\
        \midrule
        1 &  What is the average room count? & \texttt{select avg(room\_count) from properties} & 5.73\\ 
        \midrule 
        \multirow{2}{*}{2}
         & What is the average selling price? & \texttt{select avg(agreed\_selling\_price)}  & 48653794 \\
         &                                    & \texttt{from properties} & \\
        \midrule
        3 & What is the average room count? & \texttt{select avg(room\_count) from properties} & 5.73\\
        \bottomrule 
        \end{tabular}}
        \caption{\textit{Failure case: repetition.}}
        \label{tab:sub_c}
    \end{subtable}
    \caption{Examples of successful and failed interactions generated by self-play. \label{tab:gen_examples}}
\end{table*}



\subsection{Beam Search Analysis\label{sec:beam_analysis}}
\begin{figure}[t]
 \centering
 \includegraphics[width=0.4\textwidth]{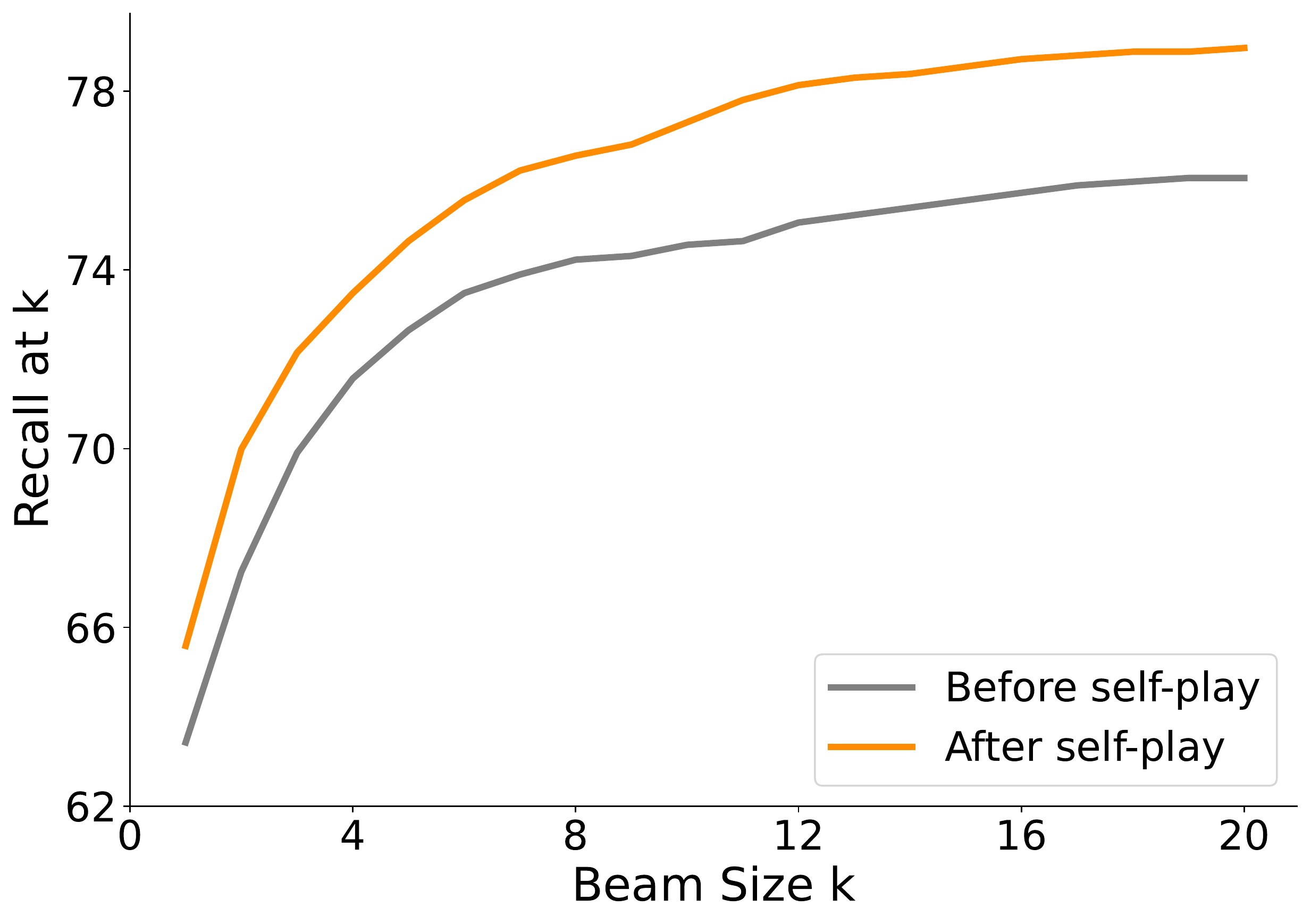}
 \vspace*{-5pt}
 \caption{Recall at k plot. Models after self-play have higher recall at different beam sizes.}
 \tao{maybe no need to put a title on the plot?} \daniella{Title removed.}
 \label{fig:recall-at-k}
\end{figure}

\daniella{Would: ``self-play increases the recall of the correct SQL" be a good hypothesis/motivation for this experiment?}
In this section we study whether self-play improves beam search by increasing the recall of the correct SQL. We first define ``Recall at beam size k'' as the probability that the ground-truth SQL is contained in the hypotheses of the beam search. This metric measures the recall of the ground-truth SQL when using the beam size k. As shown in Figure \ref{fig:recall-at-k}, we plot ``Recall at beam size k'' with k from 1 to 20. We observe that the model after self-play has higher recall compared to the model before self-play at different beam sizes. For example, the recall is 76.1 before self-play and 79.0 after self-play when k is 20. This shows that self-play can improve the recall of the ground-truth SQL. The recall at beam size 20 after self-play is 13.5 higher than the corresponding exact match score (79.0 vs.\ 65.5), demonstrating that the model has a high recall of the ground-truth SQL yet the ground-truth SQL may not have the highest beam score. 



\begin{figure}[t]
 \centering
 \includegraphics[width=0.4\textwidth]{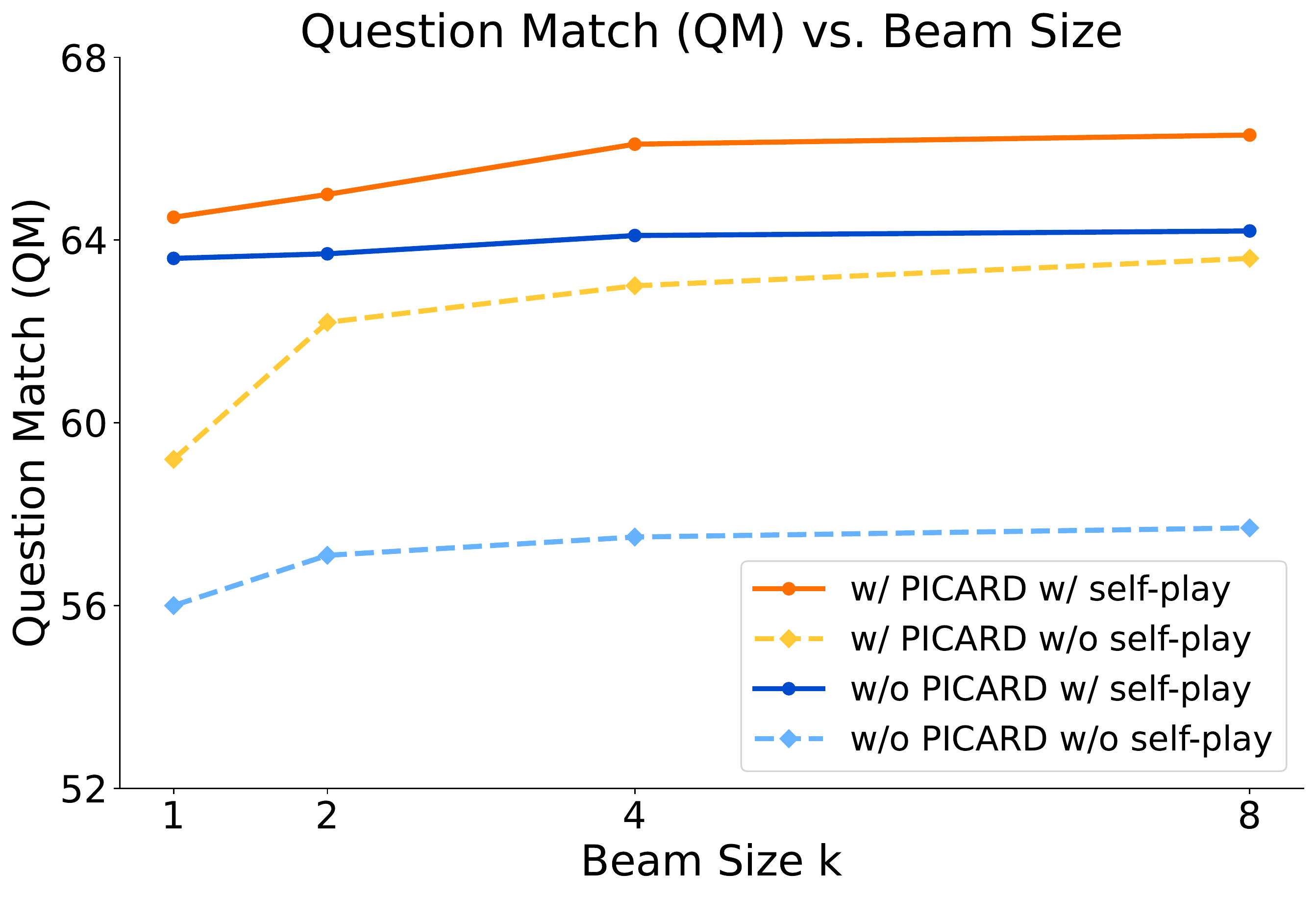}
 \vspace*{-5pt}
 \caption{Question Match with different beam sizes.}
 \label{fig:beam}
\end{figure}

As shown in Figure \ref{fig:beam}, we further plot the exact match score of the T5-Large model with 4 configurations at different beam sizes to understand the effect of beam size on model performance. In general, the exact match score improves with larger beam sizes. We observe that (1) the model with self-play outperforms the model without self-play in all configurations; (2) the models with PICARD are more sensitive to beam sizes because the performance improves significantly with increasing beam sizes;\footnote{This conforms with the plot on the Spider dataset in \citet{scholak-2021}, where the authors observe pronounced improvements with larger beam sizes.} (3) the models with self-play are less sensitive to beam sizes as they can obtain high exact match scores with even small beam sizes; (4) self-play can improve the performance with/without PICARD.



\subsection{Case Study of Self-Play Interactions\label{sec:case_analysis}}

Table \ref{tab:gen_examples} shows successful (\ref{tab:sub_a}) and failed interactions (\ref{tab:sub_b}, \ref{tab:sub_c}) generated with self-play. In Table \ref{tab:sub_a},  given the sampled goal (the same as the final system query from turn 3), the SQL-to-text model can decompose it into smaller questions over multiple turns. After asking for the locations of gas stations in the first turn, the SQL-to-text model asks for the company names of the gas stations, a theme-entity question, and then proceeds to query the assets of these companies in descending order, an answer refinement question. This demonstrates that different thematic relations are learned by the SQL-to-text model. Meanwhile, the text-to-SQL model produces the correct SQL queries in a context-dependent way.

\begin{figure}[t]
 \centering
 \includegraphics[width=0.35\textwidth]{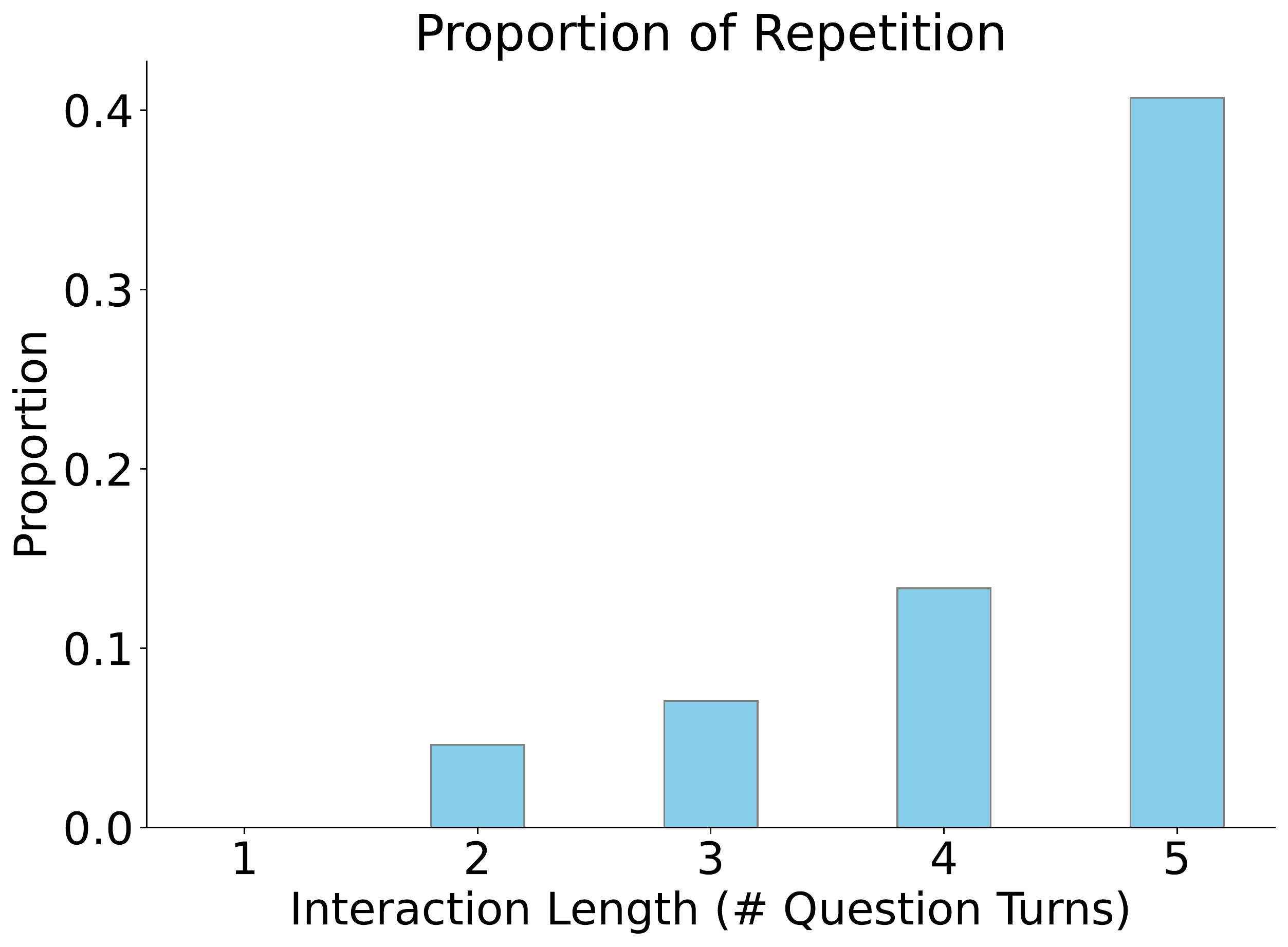}
 \vspace{-4pt}
 \caption{The proportion of repetition in generated interactions grouped by interaction length. \label{fig:repetition}}
\end{figure}

Next, we analyze the failure cases. In Table \ref{tab:sub_b}, the user utterances are not grounded to the sample query in the course of the dialogue, e.g.\ the question in the final turn does not match the semantics of the goal query. Although the final system query matches the sampled goal, language drift happens in the middle of the conversation. For example, the user utterance mentions templates in the second turn, but the text-to-SQL model ignores this keyword in the SQL query. Another failure case is the repetition of user utterances. Figure \ref{fig:repetition} shows the proportion of generated interactions in which the user utterances repeat. Repetition happens more frequently with increasing interaction lengths. Table \ref{tab:sub_c} shows an example of a repetitive interaction. Although the SQL-to-text model produces the sampled goal in the first turn, the stop-of-interaction symbol does not appear in the user utterance. As a result, the conversation continues, and the user simply repeats its first question in the third turn.


\section{Related Work}

\paragraph{Data Augmentation for Semantic Parsing}
Data augmentation \cite{feng-2021-survey} is an effective strategy to increase the diversity of training data without manually collecting new data. Data augmentation has been applied in NLP \cite{jia-2016-recombination} on various tasks such as paraphrase extraction \cite{barzilay-mckeown-2001}, machine translation \cite{sennrich-etal-2016,liu2021counterfactual}, and question-answering \cite{longpre-2019-exploration}.
Data augmentation is also widely-adopted in semantic parsing tasks \cite{jia-2016-recombination,hou-2018-augmentation, yu-2020-grappa, zhong2021grounded, wang-2021-learning,  yang-etal-2022-addressing}. Most previous work on data augmentation for text-to-SQL tasks use single-turn datasets such as SPIDER. \citet{yu-etal-2018-syntaxsqlnet} create cross-domain augmentation data with question-SQL patterns extracted from 
Spider. \citet{guo-2018} use a syntax-and-table-aware semantic parser and a copy-based latent variable model to generate SQLs and questions, respectively. \citet{wu-2021-hie} apply the abstract syntax tree grammar for SQL generation and a hierarchical SQL-to-question generation model to generate questions.

Data augmentation for multi-turn SQL-to-text datasets is under-explored. The task is more challenging compared to single-turn datasets as it requires sequential generation that takes into consideration complex contextual dependencies and user goals. \citet{zhong2021grounded} combine a forward semantic parser with a backward utterance generator to generate multi-turn interactions. The algorithm truncates the contextual window to 2 and does not condition on a global user goal during generation. As a result, this method fails to consider long-range contextual dependencies. Differently, our method conditions each turn on its full history context and the sampled user goal. This enables capturing longer contextual dependencies. For further comparison of the model performance between our proposed self-play method and \citet{zhong2021grounded}, please see Appendix \ref{app:analysis}.

 

\paragraph{Self-Play in Task-Oriented Dialogue}
As learning from real users is costly and time-consuming, self-play with user simulators has been employed in task-oriented dialogue systems \cite{levin-2000}. Two types of user simulators including rule-based \cite{schatzmann2006survey,  schatzmann2007agenda, schatzmann-2009-hidden, shah-2018, shah-etal-2018-bootstrapping}) and data-driven \cite{asri-2016,gur2018user, kreyssig-2018-neural, tseng-etal-2021-transferable} are widely-adopted. Rule-based user simulators make use of hand-crafted rules in building dialogue schedules, while data-driven user simulators are trained on gold dialogues. In task-oriented dialogue systems, each domain has its own slot-value pairs, which are domain-dependent. As a result, adapting the system to a new domain usually requires data collection, model redesigning, and retraining. Differently, SQL is domain-agnostic for text-to-SQL tasks. Self-play is well-suited for cross-domain text-to-SQL tasks as it can synthesize user and system interactions to generalize to new domains.

\paragraph{Mitigating the Exposure Bias} 
Exposure bias \cite{bengio-2015, ranzato-2015} is the mismatch between training and the generation procedure that happens when the model is only exposed to ground-truth interactions without being conditioned on its own predicted interactions. Several methods \cite{ranzato-2015,shen-etal-2016-minimum,leblond-2017, welleck-2019} have been proposed to bridge this train and test time discrepancy. Our model demonstrates the benefits of using the high-quality predicted interactions to retrain the original model and is a reasonable way to condition the model on its own prediction and mitigate the exposure bias issue.




\section{Conclusion}
We explore using self-play as a data augmentation method for generating synthetic dialogues in the cross-domain conversational semantic parsing task to address the challenge of data scarcity and cross-domain generalization. Self-play learns various thematic relations in dialogues, improves beam search, and encourages the model’s generalization to different domains. Experiments on a T5 text-to-SQL semantic parser demonstrate the benefit of our proposed method. In the future, we will study using rewards in a RL setting to guide self-play to produce better synthetic dialogues.  

\section*{Limitations}
Although the filtered dialogues after self-play are mostly grounded to the sampled user goal, some synthetic dialogues are unnatural as illustrated in section \S\ref{sec:case_analysis} Therefore, a more controlled generation of self-play that penalizes producing repetitive questions, and encourages dialogues that last longer turns would be desirable. The experiments require large GPU resources and restrict us to run self-play for one round. Running self-play with the re-trained models iteratively for multiple rounds may possibly improve the results more. Dataset-wise, in the real world, as humans do not ask questions in a controlled setting as in SParC and CoSQL, the data distribution may be more noisy and complicated. Self-play is not able to generate synthetic dialogues that diverge from the training data to simulate real-world scenarios. 



\bibliography{anthology,custom}
\bibliographystyle{acl_natbib}

\appendix

\section{Appendix}
\label{sec:appendix}

\subsection{Query Sampling Procedure in GAZP \label{app:query_sampling}}
The query sampling procedure in GAZP \cite{zhong2021grounded} is shown in Algorithm \ref{al:query_sampling}.

\begin{algorithm}[!h]
\small
\SetAlgoLined
\SetKwInOut{Input}{Input}
\SetKwInOut{Output}{Output}
\SetKw{Continue}{continue}
\SetKw{Or}{or}
\SetKw{In}{in}
\Input{Databases $\mathcal{D}$, training data $X_{tr}$. }
\Output{Sampled query $z'$.}

 Preprocess queries in training data $X_{tr}$. \\
 Replace columns and values in $X_{tr}$ with typed slots to form coarse templates $Z$. \\
 Sample a database, $d$ $\sim$ UNIFORM($\mathcal{D}$) \\
 $Z' =  \emptyset$ \\
 \For{$z$ $\in$ $Z$}{
    \If{$z$ can be filled with $d$}{$Z'$.\textit{ADD}($z$)}
}
 Build empirical distribution $P_{Z'}$ by counting their occurrences in $X_{tr}$. \\ 
 Sample a template $z' \sim P_{Z'} $. \\
 Randomly assign columns and values $s,v$ in $d$ to populate the corresponding typed slots in $z'$.  \\
 
 \Return $z'$ .
 \caption{Query Sampling Procedure in GAZP.\label{al:query_sampling}}
\end{algorithm}

\subsection{Implementation Details \label{app:implementation} }
We experiment with T5-Base (\textasciitilde220m parameters) and T5-Large (\textasciitilde770m parameters). Adam \cite{kingma-2014-adam} is used for optimization with a learning rate of 1e-4 and 5e-6 for text-to-SQL and SQL-to-text, respectively. The beam size is set to 4 for text-to-SQL and 20 for SQL-to-text.  We used 8 GPUs for all our experiments. PICARD is set to the highest mode ``Parse with Guard'', and the max number of tokens to check for PICARD is 2. Inputs longer than 512 tokens are truncated. The maximum number of self-play turns is set to 5. The threshold $w$ for filtering interactions is set to 0.5. We generate 100,000 synthetic interactions before filtering.

\subsection{Analysis on the Generated Data\label{app:analysis} }
To study the effect of the size of generated data on the final accuracy, we show the QM scores after self-play on SParC validation set with T5-base trained on different number of synthetic data before filtering. As shown in Table \ref{tab:filtersize}, we do not observe significant improvements after using 100,000 synthetic data before filtering, thus we choose the size to be 100,000 in the experiments.

\begin{table}[!h]
\centering
\scalebox{0.7}{
\begin{tabular}{llll}
\hline
  \textbf{\# of synthetic data} & \textbf{50,000} & \textbf{100,000} & \textbf{150,000}\\
\hline
  \textit{Question Match} (QM)  &  60.2 &  62.4  & 62.5  \\
\hline
\end{tabular}}
\caption{The 
QM score after self-play on SParC validation set with T5-base trained on different number of synthetic data before filtering. 
\label{tab:filtersize}
}
\end{table}

In our experiments, filtering is applied to discard low-quality synthetic interactions that diverge from the user’s goals. We find that training on low-quality interactions gives negative effects for the final performance. We study the effect of changing the filter threshold value $w$, as shown in Table \ref{tab:threshold}. The final threshold $w$ for filtering interactions is set to 0.5 as a larger threshold aggressively filters most synthetic dialogues that are of hard/extra-hard difficulties.

\begin{table}[!h]
\centering
\scalebox{0.7}{
\begin{tabular}{lllll}
\hline
  \textbf{$w$} & \textbf{0} & \textbf{0.3} & \textbf{0.5} & \textbf{0.7} \\
\hline
  \textit{Question Match} (QM)  &  56.2 &  60.5  &  62.4 & 61.8  \\
\hline
\end{tabular}}
\caption{The 
QM score on SParC validation set with T5-base trained with different filter value $w$. 
\label{tab:threshold}
}
\end{table}

We further study if conditioning on a user goal $\mathcal{G}$ when generating interactions is necessary. When we ablate the user goal, the QM score on SParC drops from 62.4 to 59.8. We argue that it is important to condition on the user goal to obtain grounded interactions.

We also reimplement the method used in \citet{zhong2021grounded} by ablating both the user goal and the context. The QM score on SParC drops from 62.4 to 58.3. We argue that it is important to condition on the user goal and the full context to obtain grounded interactions.

\subsection{Template Examples \label{app:more_templates}}

\begin{table}[h]
\centering
\scalebox{0.67}{
\begin{tabular}{ll}
\hline
    \textbf{Top Templates Unseen in Train} & \textbf{Proportion}   \\
\hline
 \texttt{select text\_col\_0 group\_by key\_col\_0}   &  0.19\%   \\
 \texttt{order\_by count (*\_col\_0) desc limit\_value}   &    \\
 \midrule{}
 \texttt{select number\_col\_0 group\_by number\_col\_0 } & 0.12\%  \\
 \texttt{order\_by count (*) desc limit\_value} &  \\
 \midrule{}
 \texttt{select text\_col\_0 , count (*\_col\_0)}  & 0.07\%   \\
 \texttt{group\_by text\_col\_0 order\_by count (*) desc}  &    \\
\hline
\end{tabular}}
\caption{Examples of generated templates unseen in SParC train.}
\end{table}

\begin{table*}[h] 
    \centering
    \scalebox{0.8}{
    \begin{tabular}{lll}
        \toprule
        \textbf{Template and Example} & \textbf{Improvement} & \\
        \midrule
        \texttt{select sum (number\_col\_0)} & 50  \\
        e.g. \texttt{SELECT sum(number\_products) FROM shop} &  \\
        \midrule
        \texttt{select text\_col\_0 , count (*\_col\_0) group\_by text\_col\_0} & 42.9  \\
        e.g.  \texttt{SELECT Nationality ,  COUNT(*) FROM people GROUP BY Nationality} &  \\
        \midrule
        \texttt{select text\_col\_0 where key\_col\_0 = value} & 28.6  \\ 
        e.g. \texttt{SELECT AirportName FROM AIRPORTS WHERE AirportCode  =  "AKO"} &  \\
        \midrule 
        \texttt{select text\_col\_0 where key\_col\_0 not in (select key\_col\_1)} & -22.2 \\
        e.g. \texttt{SELECT Name FROM people WHERE People\_ID NOT IN (SELECT People\_ID FROM poker\_player)} & \\ 

    \bottomrule
    \end{tabular}
    }
    \caption{Examples of templates on which self-play improves (or reduces) performance. \label{tab:example_improved}}
\end{table*}

\end{document}